\documentclass[11pt,a4paper]{article}
\usepackage[hyperref]{acl2020}
\usepackage{times}
\usepackage{latexsym}

\usepackage{microtype}

\usepackage{color}
\usepackage{multirow}

\usepackage{verbatim}
\usepackage{booktabs}
\usepackage{hhline}

\usepackage{footnote}
\usepackage{footmisc}
\usepackage{threeparttable}

\usepackage{amsmath,amssymb,bm}
\usepackage{algorithm}
\usepackage{algpseudocode}

\usepackage{graphicx}
\usepackage{subfigure}
\usepackage{url}

\aclfinalcopy 

\title{MixText: Linguistically-Informed Interpolation of Hidden Space for Semi-Supervised Text Classification}

\author{Jiaao Chen \\
  Georgia Tech \\
  \texttt{jchen896@gatech.edu} \\\And
  Zichao Yang \\
  CMU \\
  \texttt{zichaoy@cs.cmu.edu} \\ \And
  Diyi Yang \\
  Georgia Tech \\
  \texttt{dyang888@gatech.edu} \\
  }

\date{}

\begin{document}
\maketitle
\begin{abstract}

This paper presents MixText, a semi-supervised learning method for text classification, which uses our newly designed data augmentation method called TMix. TMix creates a large amount of augmented training samples by interpolating text in {\em hidden space}. Moreover, we leverage recent advances in data augmentation to guess low-entropy labels for unlabeled data, hence making them as easy to use as labeled data.
By mixing labeled, unlabeled and augmented data, MixText significantly outperformed current pre-trained and fined-tuned 
models and other state-of-the-art semi-supervised learning methods on several text classification benchmarks. 
The improvement is especially prominent when supervision is extremely limited. We have publicly released our
code at \url{https://github.com/GT-SALT/MixText}.

\end{abstract}

\section{Introduction}
In the era of deep learning, research has achieved extremely good performance in most 
supervised learning settings  \cite{0483bd9444a348c8b59d54a190839ec9,yang2016hierarchical}. 
However, when there is only limited labeled data, supervised deep learning models often suffer from over-fitting \cite{xie2019unsupervised}. 
This strong dependence on labeled data largely prevents neural network models from being applied to new settings or real-world situations due to the need of large amount of time, money, and expertise to obtain enough labeled data. 
As a result, semi-supervised learning has received much attention to utilize both labeled and unlabeled data for different learning tasks, as unlabeled data is always much easier and cheaper to collect \cite{abs-1109-2047}.

\begin{figure}[th!]
\centering
\includegraphics[width=0.5\textwidth]{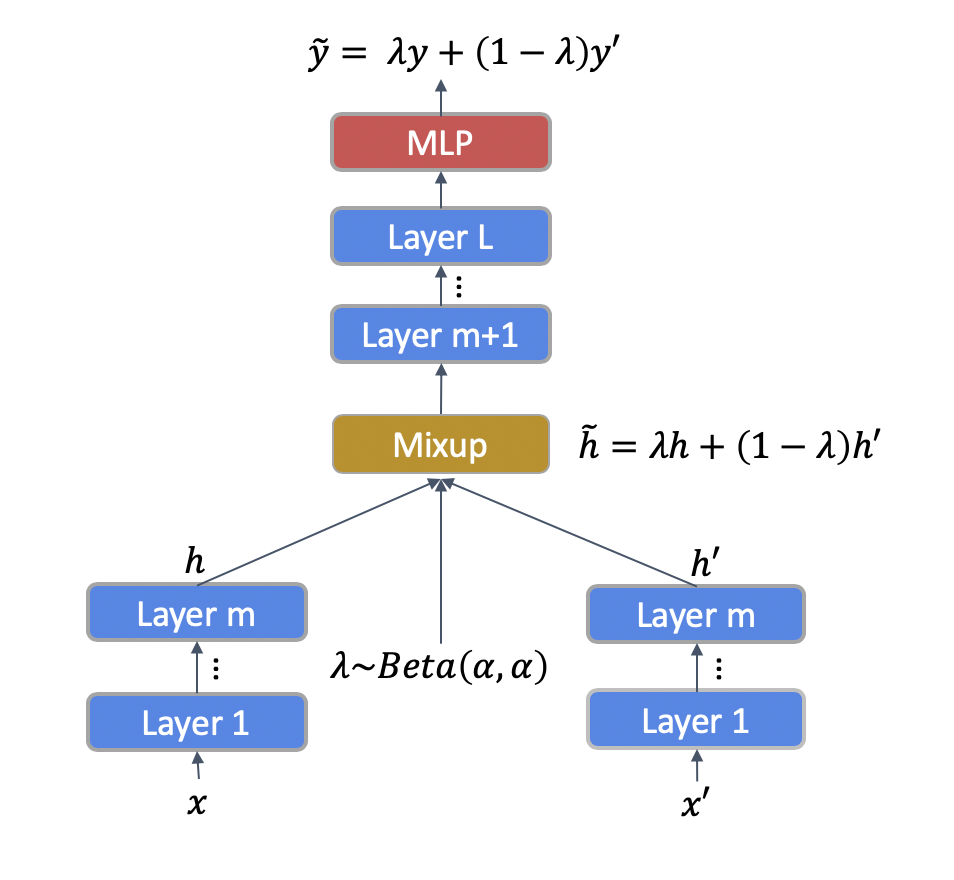} 
\caption[]{TMix takes in two text samples $x$ and $x'$ with labels $y$ and $y'$, mixes their hidden states $h$ and $h'$ at layer $m$ 
with weight $\lambda$ into $\tilde{h}$, and then continues forward passing to predict the mixed labels $\tilde{y}$.}
\label{fig:tex_mixup}
\end{figure}
This work takes a closer look at semi-supervised text classification, one of the most fundamental tasks in language technology communities. 
Prior research on semi-supervised text classification can be categorized into several classes: (1) utilizing variational auto encoders (VAEs) to reconstruct the sentences and predicting sentence labels with latent variables learned from reconstruction such as \cite{mchen-variational-18, YangHSB17, abs-1906-02242}; (2) encouraging models to output confident predictions on unlabeled data for self-training like \cite{article:Pseudo-Label, Grandvalet:2004:SLE:2976040.2976107, Meng:2018:WNT:3269206.3271737};
(3) performing consistency training after adding adversarial noise \cite{MiyatoMKI19, shen2018deep} or data augmentations \cite{xie2019unsupervised};
(4) large scale pretraining with unlabeld data, then finetuning with labeled
data~\cite{devlin-etal-2019-bert}.
Despite the huge success of those models, most prior work utilized labeled and unlabeled data {\em separately} in a way that no supervision can transit from labeled to unlabeled data or from unlabeled to labeled data. 
As a result, most semi-supervised models can easily still overfit on the very limited labeled data, despite unlabeled data is abundant.

To overcome the limitations, in this work, we introduce a new data augmentation method, called \textbf{TMix} (Section~\ref{section3}),  inspired by the recent success of Mixup \cite{abs-1906-02242, abs-1905-02249} on image classifications.
TMix, as shown in Figure~\ref{fig:tex_mixup}, takes in two text instances, 
and interpolates them in their corresponding hidden space. 
Since the combination is continuous, TMix has the potential to create infinite mount of new augmented data samples, thus can drastically avoid overfitting. 
Based on TMix, we then introduce a new semi-supervised learning method for text classification called \textbf{MixText} (Section~\ref{section4}) to explicitly model the relationships between {\em labeled and unlabeled} samples, thus overcoming the limitations of
previous semi-supervised models stated above. In a nutshell, MixText first guesses low-entropy labels for unlabeled data, then uses TMix to interpolate the label and unlabeled data. MixText can facilitate mining implicit relations between sentences by encouraging models to behave linearly in-between training examples, and utilize information from unlabeled sentences while learning on labeled sentences. In the meanwhile, MixText exploits several semi-supervised learning techniques to further utilize unlabeled data including self-target-prediction \cite{journals/corr/LaineA16}, entropy minimization \cite{Grandvalet:2004:SLE:2976040.2976107}, and consistency regularization \cite{abs-1905-02249, xie2019unsupervised} after back translations.

To demonstrate the effectiveness of our method, we conducted experiments (Section~\ref{section5}) on four benchmark text classification datasets and compared our method with previous state-of-the-art semi-supervised method, including those built upon models pre-trained with large amount of unlabeled data, in terms of accuracy on test sets. 
We further performed ablation studies to demonstrate each component's influence on models' final performance. Results show that our MixText method significantly outperforms baselines especially when the given labeled training data is extremely limited.

\section{Related Work}
\subsection{Pre-training and Fine-tuning Framework}
The pre-training and fine-tuning framework has achieved huge success on NLP applications in recent years, and has been applied to a variety of NLP tasks \cite{noauthororeditor,conf/aaai/ChenCY19, akbik-etal-2019-pooled}. \citet{howard-ruder-2018-universal} proposed to pre-train a language model on a large general-domain corpus and fine-tune it on the target task using some novel techniques like discriminative fine-tuning,
slanted triangular learning rates, and gradual unfreezing. In this manner, such pre-trained models show excellent performance even with small amounts of labeled data. Pre-training methods are often designed with different objectives such as language modeling \cite{peters-etal-2018-deep, howard-ruder-2018-universal, abs-1906-08237} and masked language modeling \cite{devlin-etal-2019-bert, abs-1901-07291}. Their performances are also improved with training larger models on more data \cite{abs-1906-08237, abs-1907-11692}. 

\subsection{Semi-Supervised Learning on Text Data}
Semi-supervised learning has received much attention in the NLP community \cite{abs-1906-02242,clark2018semi,yang2015weakly}, as unlabeled data is often  plentiful compared to labeled data. For instance, \citet{abs-1906-02242, mchen-variational-18, YangHSB17} leveraged variational auto encoders (VAEs) in a form of sequence-to-sequence modeling on text classification and sequential labeling. \citet{shen2018deep} utilized adversarial and virtual adversarial training to the text domain by applying perturbations to the word embeddings. \citet{yang2019let} took advantage of  hierarchy structures to utilize supervision from higher level labels to lower level labels. \citet{xie2019unsupervised} exploited consistency regularization on unlabeled data after back translations and tf-idf word replacements. \citet{clark2018semi} proposed cross-veiw training for unlabeled data, where they used an auxiliary prediction modules that see restricted views of the input (e.g., only part of a sentence) and match the predictions of the full model seeing the whole input. 

\subsection{Interpolation-based Regularizers}
Interpolation-based regularizers (e.g., Mixup) have been recently proposed for 
supervised learning \cite{abs-1710-09412, pmlr-v97-verma19a} and semi-supervised learning \cite{abs-1905-02249, ijcai2019-504} for image-format data by overlaying two input images and combining image labels as virtual training data and have achieved state-of-the-art performances across a variety of tasks like image classification and network architectures. 
Different variants of mixing methods have also been designed  such as performing interpolations in the input space \cite{abs-1710-09412}, combining interpolations and cutoff \cite{abs-1905-04899}, and doing interpolations in the hidden space representations \cite{pmlr-v97-verma19a,Verma2019GraphMixRT}. However, such interpolation techniques have not been explored in the NLP field because most input space in text is discrete, i.e., one-hot vectors instead of continues RGB values in images, and text is generally more complex in structures.

\subsection{Data Augmentations for Text}
When labeled data is limited, data augmentation has been a useful technique to increase the amount of training data. For instance,  in computer vision, images are shifted, zoomed in/out, rotated, flipped, distorted, or shaded with a hue \cite{abs-1712-04621} for training data augmentation. But it  is relatively challenging  to augment text data because of its complex syntactic and semantic structures. Recently, \citet{abs-1901-11196} utilized synonym replacement, random insertion, random swap and random deletion for text data augmentation. Similarly, \citet{kumar-etal-2019-submodular} proposed a new paraphrasing formulation in terms of
monotone sub-modular function maximization to obtain highly diverse paraphrases, and \citet{xie2019unsupervised} and \citet{workshop_aaai_jiaao} applied back translations \cite{SennrichHB15a} and word replacement to generate paraphrases on unlabeled data for consistency training. Other work which also investigates noise and its incorporation into semi-supervised named entity classification \cite{lakshmi-narayan-etal-2019-exploration, nagesh2018exploration}.

\section{TMix} \label{section3}
In this section, we extend Mixup--a data augmentation method originally proposed
by ~\cite{abs-1710-09412} for images--to text modeling. 
The main idea of Mixup is very simple: given
two labeled data points $(\mathbf{x}_i, \mathbf{y}_i)$ and $(\mathbf{x}_j, \mathbf{y}_j)$,
where $\mathbf{x}$ can be an image and $\mathbf{y}$ is the one-hot representation of the label,
the algorithm creates virtual training samples by linear interpolations:
\begin{align}
\tilde{\mathbf{x}} = \text{mix}(\mathbf{x}_i, \mathbf{x}_j) =& \lambda \mathbf{x}_i + (1-\lambda)\mathbf{x}_j, \label{eq:mix_x} \\
\tilde{\mathbf{y}} = \text{mix}(\mathbf{y}_i, \mathbf{y}_j) =& \lambda \mathbf{y}_i + (1-\lambda)\mathbf{y}_j, \label{eq:mix_y}
\end{align}
where $\lambda \in [0, 1]$. The new virtual training
samples are used to train a neural network model. Mixup can be interpreted in different ways.
On one hand, Mixup can be viewed a data augmentation 
approach which creates new data samples based on the original training set. On
the other hand, it enforces a regularization on the model to
behave linearly among the training data.
Mixup was demonstrated to work well on continuous image data~\cite{abs-1710-09412}. 
However, extending it to text seems challenging since 
it is infeasible to compute the interpolation of discrete tokens. 

To this end, we propose a novel method to overcome this challenge --- 
{\em interpolation in textual hidden space}. Given a sentence, we often
use a multi-layer model like BERT~\cite{devlin-etal-2019-bert}
to encode the sentences to get the semantic representations, based on which
final predictions are made. Some prior work~\cite{BowmanVVDJB16}
has shown that decoding from an interpolation of two hidden vectors
generates a new sentence with mixed meaning of two original sentences. 
Motivated by this, we propose to apply
interpolations within hidden space as a data augment method for text. 
For an encoder with $L$ layers, we choose to mixup the hidden
representation at the $m$-th layer, $m\in[0, L]$.

As demonstrated in 
Figure~\ref{fig:tex_mixup}, 
we first compute the hidden representations of two text samples
separately in the bottom layers.
Then we mix up the hidden representations at layer $m$, and feed the 
interpolated hidden representations to the upper layers.
Mathematically, denote the $l$-th layer in the encoder network as 
${g_l(.; \bm{\theta})}$, hence the hidden representation of the 
$l$-th layer can be computed as
$\mathbf{h}_l = g_l(\mathbf{h}_{l-1}; \bm{\theta}).$
For two text samples $\mathbf{x}_i$ and $\mathbf{x}_j$, 
define the $0$-th layer as the embedding layer, i.e., $\mathbf{h}_0^i = \mathbf{W}_{\text{E}}\mathbf{x}_i, 
\mathbf{h}_0^j=\mathbf{W}_{\text{E}}\mathbf{x}_j$,
then the hidden representations of the two samples from the lower layers are:
\begin{align*}
\mathbf{h}_l^i =& g_l(\mathbf{h}_{l-1}^i; \bm{\theta}), l\in[1, m],\\
\mathbf{h}_l^j =& g_l(\mathbf{h}_{l-1}^j; \bm{\theta}), l\in[1, m].
\end{align*}

The mixup at the $m$-th layer and continuing forward passing to upper layers are defined as:
\begin{align*}
&\tilde{\mathbf{h}}_m =  \lambda \mathbf{h}_m^i + (1-\lambda) \mathbf{h}_m^j,\\
&\tilde{\mathbf{h}}_l =  g_l(\tilde{\mathbf{h}}_{l-1}; \bm{\theta}), l \in[m+1, L].
\end{align*}
We call the above method {\bf TMix} and define the new mixup operation
as the whole process to get $\tilde{\mathbf{h}}_L$:
\begin{align*}
    \text{TMix}(\mathbf{x}_i, \mathbf{x}_j; g(.;\bm{\theta}), \lambda, m) = \tilde{\mathbf{h}}_L.
\end{align*}

By using an encoder model $g(.;\bm{\theta})$, TMix interpolates textual semantic hidden 
representations as a type of data augmentation. In contrast with Mixup defined
in the data space in Equation~\ref{eq:mix_x}, TMix depends on an encoder function, hence defines a much broader
scope for computing interpolations. For ease of notation,
we drop the explicit dependence on $g(.; \bm{\theta})$, $\lambda$ and $m$ in 
notations and denote it simply as 
$\text{TMix}(\mathbf{x}_i, \mathbf{x}_j)$ in the following sections.

In our experiments, we sample the mix parameter $\lambda$ from a 
Beta distribution for every batch to perform the interpolation :
\begin{align*}
    \lambda &\sim \text{Beta}(\alpha , \alpha),\\
    \lambda &= \text{max}(\lambda, 1-\lambda),
\end{align*}  
in which $\alpha$ is the hyper-parameter to control the distribution of $\lambda$.
In TMix, we mix the labels in the same way as Equation~\ref{eq:mix_y} and
then use the pairs $(\tilde{\mathbf{h}}_L, \tilde{y})$ as inputs
for downstream applications.

Instead of performing mixup at random input layers like \citet{pmlr-v97-verma19a}, choosing which layer of the hidden representations to mixup is an 
interesting question to investigate.
In our experiments, we use 12-layer 
BERT-base~\cite{devlin-etal-2019-bert} as our encoder model.
Recent work~\cite{jawahar-etal-2019-bert} has studied what BERT 
learned at different layers. Specifically, the authors found 
\{3,4,5,6,7,9,12\} layers have the most representation power in BERT 
and each layer captures different types of information ranging from surface, 
syntactic to semantic level representation of text. For instance, the 9-th layer has predictive power in semantic tasks like checking  random swapping of coordinated clausal conjuncts, while the 3-rd layer performs best in surface tasks like predicting sentence length.
 
Building on those findings, we choose the layers that contain both syntactic and 
semantic information as our mixing layers, namely $\mathbf{M} = \{7, 9, 12\}$.
For every batch, we {\em randomly sample} $m$, the layer to mixup 
representations, from the set $\mathbf{M}$ computing the interpolation. We also performed ablation study in 
Section~\ref{ablation} to show how TMix's performance changes
with different choice of mix layer sets. 

\paragraph{Text classification} Note that TMix provides a general
approach to augment text data, hence
can be applied to any downstream tasks. In this paper, we focus on text
classification and leave other applications as potential future work.
In text classification, we minimize the KL-divergence between
the mixed labels and the probability from the classifier as the supervision loss:
\begin{align*}
L_\text{TMix} = \text{KL}( \text{mix}(\mathbf{y}_i, \mathbf{y}_j) || p(\text{TMix}(\mathbf{x}_i, \mathbf{x}_j); \bm{\phi})
\end{align*}
where $p(.;\bm{\phi})$ is a classifier on top of the encoder model. 
In our experiments, we implement the classifier as a two-layer MLP, which takes
the mixed representation $\text{TMix}(\mathbf{x}_i, \mathbf{x}_j)$ as input and returns
a probability vector. We jointly optimize over the encoder parameters $\bm{\theta}$ and the
classifier parameters $\bm{\phi}$ to train the whole model.

\begin{figure*}[t!]
\centering
\includegraphics[width=1.0\textwidth]{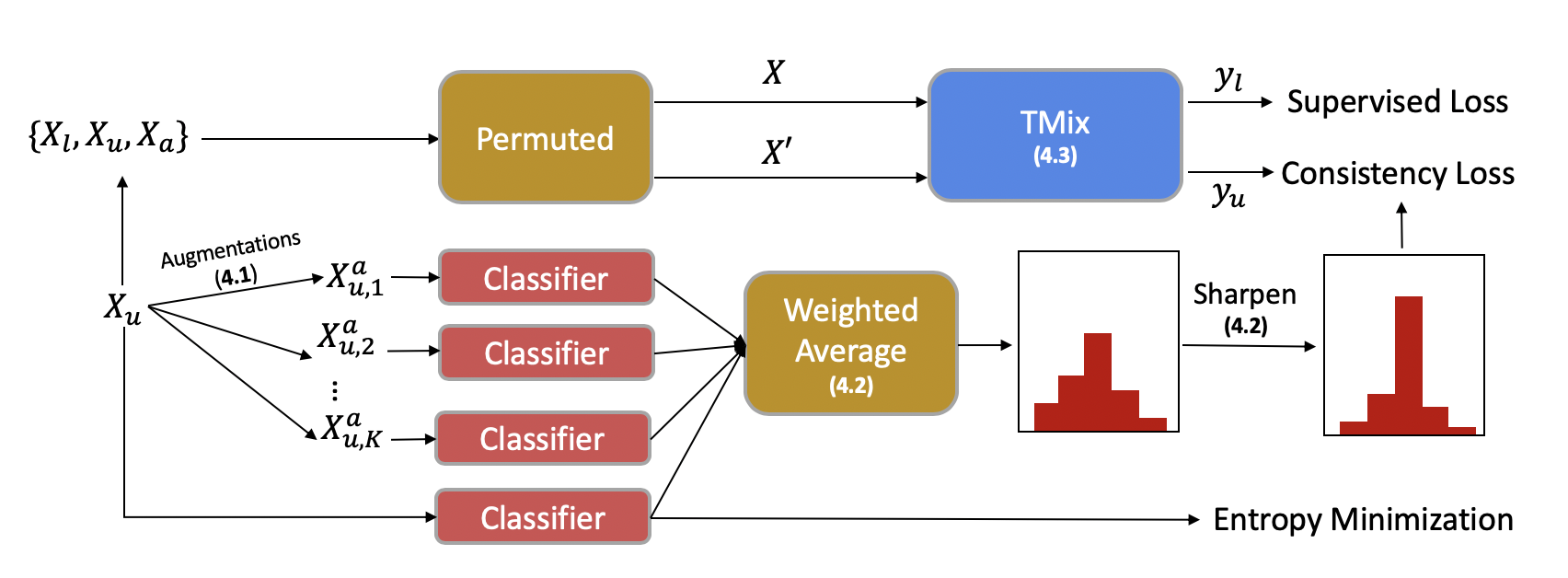}
\caption[]{Overall Architecture of MixText.  MixText takes in labeled data and unlabeled data, conducts augmentations and predicts labels for unlabeled data, performs TMix over labeled and unlabeled data, and computes supervised loss, consistency loss and entropy minimization term.}%
\label{fig:overall_model}
\end{figure*}

\section{Semi-supervised MixText} \label{section4}
In this section, we demonstrate how to utilize the TMix to help 
semi-supervised learning. Given a limited labeled text set
$\mathbf{X}_l = \{\mathbf{x}^l_1, ..., \mathbf{x}^l_n\}$, with their labels
$\mathbf{Y}_l = \{\mathbf{y}^l_1, ..., \mathbf{y}^l_n\}$ and a large 
unlabeled set $\mathbf{X}_u = \{\mathbf{x}^u_1, ..., \mathbf{x}^u_m\}$, 
where $n$ and $m$ are the number of data points in each set. $\mathbf{y}_i^l \in \{0,1\}^C$
is a one-hot vector and $C$ is the number of classes. Our goal 
is to learn a classifier that efficiently utilizes both labeled data and unlabeled data.

We propose a new text semi-supervised learning framework called {\bf MixText} 
\footnote{Note that MixText is a semi-supervised learning framework while TMix
is a data augmentation approach.}. The core idea behind our framework is to leverage TMix
both on labeled and unlabeled data for semi-supervised learning. To fulfill 
this goal, we come up a label guessing method to generate labels for the 
unlabeled data in the training process. With the guessed labels,
we can treat the unlabeled data as additional labeled data and perform TMix
for training. Moreover, we combine TMix with additional data augmentation
techniques to generate large amount of augmented data, which is a key component
that makes our algorithm work well in setting with extremely limited supervision.
Finally, we introduce an entropy minimization loss that encourages
the model to assign sharp probabilities on unlabeled data samples, which further
helps to boost performance when the number of classes $C$ is large.
The overall architecture is shown in Figure~\ref{fig:overall_model}. We will explain each 
component in detail.

\subsection{Data Augmentation}
Back translations~\cite{abs-1808-09381} is a common data augmentation technique and
can generate diverse paraphrases while preserving the 
semantics of the original sentences.
We utilize back translations to paraphrase the unlabeled data. 
For each $\mathbf{x}_i^u$ in the unlabeled text set $\mathbf{X}_u$, we generate 
$K$ augmentations ${\mathbf{x}}_{i,k}^a = \text{augment}_k(\mathbf{x}_i^u), k \in [1,K]$ by 
back translations with different intermediate languages. For example, we can 
translate original sentences from English to German and then translate them back to get the paraphrases.
In the augmented text generation, we employ random sampling with a tunable 
temperature instead of beam search to ensure the diversity. 
The augmentations are then used for generating labels for the unlabeled data, which
we describe below.

\subsection{Label Guessing}
For an unlabeled data sample $\mathbf{x}_i^u$ and its $K$ augmentations 
${\mathbf{x}}_{i,k}^a$, we generate the label for them using
weighted average of the predicted results from the current model:
\begin{align*}
    \mathbf{y}^u_i = \frac{1}{w_{ori} + \sum_k w_k }( w_{ori}p(\mathbf{x}^u_i) + \sum_{k=1}^K w_k p({\mathbf{x}}^a_{i,k}))).
\end{align*}
Note that $\mathbf{y}^u_i$ is a probability vector. We expect the model
to predict consistent labels for different augmentations.
Hence, to enforce the constraint, we use the weighted average of 
all predictions, rather than the prediction of any single data sample,
as the generated label. 
Moreover, by explicitly introducing the weight $w_{ori}$ and $w_k$,
we can control the contributions of different quality of augmentations to
the generated labels. Our label guessing method improves over 
\cite{TarvainenV17} which utilizes 
teacher and student models to predict labels for unlabeled data, and UDA
\cite{xie2019unsupervised} that just uses $p(\mathbf{x}^u_i)$ as generated
labels.

To avoid the weighted average being too uniform, we 
utilize a sharpening function over predicted labels. 
Given a temperature hyper-parameter $T$:
\begin{align*}
\text{Sharpen}(\mathbf{y}^u_i, T) = \frac{(\mathbf{y}^u_i)^{\frac{1}{T}}}{||(\mathbf{y}_i^u)^{\frac{1}{T}}||_1},
\end{align*}
where $||.||_1$ is $l_1$-norm of the vector. When $T\to 0$, the generated
label becomes a one-hot vector.

\subsection{TMix on Labeled and Unlabeled Data}
After getting the labels for unlabeled data, we merge the labeled text $\mathbf{X}_l$,
unlabeled text $\mathbf{X}_u$ and unlabeled augmentation 
text ${\mathbf{X}}_{a}=\{x^a_{i,k}\}$ together
to form a super set $\mathbf{X} = \mathbf{X}_l \cup \mathbf{X}_u \cup {\mathbf{X}}_{a}$.
The corresponding labels are $\mathbf{Y} = \mathbf{Y}_l \cup \mathbf{Y}_u \cup \mathbf{Y}_a$,
where $\mathbf{Y}^a = \{\mathbf{y}^{a}_{i,k}\}$ and we define $\mathbf{y}^a_{i, k} = 
\mathbf{y}^u_i$, i.e., the all augmented samples share the same generated label
as the original unlabeled sample. 

In training, we randomly sample two data points $\mathbf{x}, \mathbf{x}^\prime \in \mathbf{X}$,
then we compute $\text{TMix}(\mathbf{x}, \mathbf{x}^\prime)$, $\text{mix}(\mathbf{y}, \mathbf{y}^\prime)$
and use the KL-divergence as the loss:
\begin{align*}
L_\text{TMix} = \mathbb{E}_{\mathbf{x}, \mathbf{x}^\prime \in \mathbf{X}} \text{KL}(\text{mix}(\mathbf{y}, \mathbf{y}^\prime) || p(\text{TMix}(\mathbf{x}, \mathbf{x}^\prime)).
\end{align*}

Since $\mathbf{x}, \mathbf{x}^\prime$ are randomly sampled from $\mathbf{X}$, 
we interpolate text from many different categories: mixup among 
among labeled data, mixup of labeled and unlabeled data and mixup of unlabeled data. 
Based on the categories of the samples, the loss can be divided into two types:
\paragraph{Supervised loss}
When $\mathbf{x} \in \mathbf{X}_l$, the majority information we are actually using is from the labeled data, hence training 
the model with supervised loss.

\paragraph{Consistency loss} When the samples are from unlabeled or augmentation
set, i.e., $\mathbf{x} \in \mathbf{X}^u \cup \mathbf{X}^a$, most information coming from unlabeled data, 
the KL-divergence is a type of consistency loss, constraining
augmented samples to have the same labels with the original data sample.

\subsection{Entropy Minimization}
To encourage the model to produce confident labels on unlabeled data,
we propose to minimize the entropy of prediction probability on
unlabeled data as a self-training loss:
\begin{equation*} \label{entropy minimization}
    L_\text{margin} = \mathbb{E}_{\mathbf{x}\in \mathbf{X}_u} \text{max}(0, \gamma - ||\mathbf{y}^u||_2^2),
\end{equation*}
where $\gamma$ is the margin hyper-parameter. We minimize the entropy of the probability
vector if it is larger than $\gamma$.

Combining the two losses, we get the overall objective function of MixText:
\begin{align*}
    L_\text{MixText} = L_\text{TMix} + \gamma_m L_\text{margin}.
\end{align*}

\section{Experiments} \label{section5}
\subsection{Dataset and Pre-processing}
We performed experiment with four English text classification benchmark datasets: AG News \cite{ZhangZL15}, BPpedia \cite{lrec12mendes2}, Yahoo! Answers \cite{Chang:2008:ISR:1620163.1620201} and IMDB \cite{Maas:2011:LWV:2002472.2002491}. We used the original test set as our test set and randomly sampled from the training set to form the training unlabeled set and development set. The dataset statistics and split information are presented in Table~\ref{dataset}.

For unlabeled data, we selected German and Russian as intermediate languages for back translations using FairSeq\footnote{\url{https://github.com/pytorch/fairseq}}, and the random sampling temperature was 0.9. Here is an example, for a news from AG News dataset: \textsl{``Oil prices rallied to a record high  above  \$55 a barrel on Friday on rising fears of a winter fuel  supply crunch and robust economic growth in China, the world's  number two user"}, the augment texts through German and Russian are: \textsl{``Oil prices surged to a record high above \$55 a barrel on Friday on growing fears of a winter slump and robust economic growth in world No.2 China"} and \textsl{``Oil prices soared to record highs above \$55 per barrel on Friday amid growing fears over a winter reduction in U.S. oil inventories and robust economic growth in China, the world's second-biggest oil consumer"}.

\begin{table*}[ht]
\centering
\begin{tabular}{cccccc} 
\hline
\textbf{Dataset} & \textbf{Label Type} &\textbf{Classes}  &\textbf{Unlabeled} &\textbf{Dev}  &\textbf{Test} \\ \hline 
AG News &News Topic &4 &5000 &2000 &1900 \\
DBpedia &Wikipeida Topic&14 &5000 &2000 &5000 \\
Yahoo! Answer &QA Topic  &10 &5000 &5000 &6000 \\
IMDB &Review Sentiment &2 &5000 &2000 &12500 \\
\hline 
\end{tabular}\caption{Dataset statistics and dataset split. The number of unlabeled data, dev data and test data in the table means the number of data per class.}\label{dataset}
\end{table*}

\begin{table*}[ht]
\centering
\begin{tabular}{|c||c|c|c|c||c||c|c|c|c|c|c|c|c|}
\hline
\textbf{Datset} &\textbf{Model}& \textbf{10} &\textbf{200} & \textbf{2500} &\textbf{Dataset} &\textbf{Model} & \textbf{10} &\textbf{200} & \textbf{2500}    \\ \hline 

 \multirow{4}{*}{AG News}    & VAMPIRE & - &83.9  & 86.2 &\multirow{4}{*}{DBpedia} & VAMPIRE&- &-  &- \\ \cline{2-5}\cline{7-10}
  & BERT      &  69.5 & 87.5  &90.8 &{} &BERT &95.2 &98.5 &99.0   \\ \cline{2-5}\cline{7-10}
   & TMix*     & 74.1 &88.1  &  91.0    &{} &TMix* &96.8 &98.7  &99.0       \\ \cline{2-5} \cline{7-10}
   & UDA     & {84.4} &{88.3}  &  {91.2}    &{} &UDA &{97.8} &{98.8}  &{99.1}      \\ \cline{2-5}\cline{7-10}
     &   MixText*   &\textbf{88.4} & \textbf{89.2} & \textbf{91.5}  &{} &MixText* &\textbf{98.5} &\textbf{98.9}  &\textbf{99.2} \\ \hline \hline

 \multirow{4}{*}{Yahoo!}    & VAMPIRE & - &59.9  & 70.2 &\multirow{4}{*}{IMDB} & VAMPIRE&- &82.2  &85.8 \\ \cline{2-5}\cline{7-10}
   & BERT      &  56.2 & 69.3  &73.2 &{} &BERT &67.5 &86.9 &89.8   \\\cline{2-5}\cline{7-10} 
   & TMix*     & 58.6 &69.8  &  73.5    &{} &TMix* &69.3 &87.4  &90.3       \\ \cline{2-5}\cline{7-10}
    & UDA    & {63.2} &{70.2}  &  {73.6}    &{} &UDA &{78.2} &89.1  &{90.8}      \\ \cline{2-5}\cline{7-10}
  &   MixText*   &\textbf{67.6} & \textbf{71.3} & \textbf{74.1}  &{} &MixText* &\textbf{78.7} &\textbf{89.4}  &\textbf{91.3} \\ \hline

\end{tabular}\caption{Performance (test accuracy(\%)) comparison with baselines. The results are averaged after three runs to show the significance \cite{dror-etal-2018-hitchhikers}, each run takes around 5 hours. Models are trained with 10, 200, 2500 labeled data per class. VAMPIRE, Bert, and TMix do not use unlabeled data during training while UDA and MixText utilize unlabeled data. * means our models. }\label{tab:main_result}
\end{table*}

\subsection{Baselines}
To test the effectiveness of our method, we compared it with several recent models:
\begin{itemize}
    \item \textbf{VAMPIRE} \cite{abs-1906-02242}: VAriational Methods for Pretraining In Resource-limited Environments(VAMPIRE) pretrained a unigram document model as a variational autoencoder on in-domain, unlabeled data and used its internal states as features in a downstream classifier.
    \item \textbf{BERT} \cite{devlin-etal-2019-bert}: We used the pre-trained BERT-based-uncased model\footnote{\url{https://pypi.org/project/pytorch-transformers/}} and fine-tuned it for the classification. In details, we used average pooling over the output of BERT encoder and the same two-layer MLP as used in MixText to predict the labels. 
    \item \textbf{UDA} \cite{xie2019unsupervised}: Since we do not have access to TPU and need to use smaller amount of unlabeled data, we implemented Unsupervised Data Augmentation(UDA) using pytorch by ourselves. Specifically, we used the same BERT-based-uncased model, unlabeled augment data and batch size as our MixText, used original unlabeled data to predict the labels with the same softmax sharpen temperature as our MixText and computed consistency loss between augmented unlabeled data.
\end{itemize}

\subsection{Model Settings}
We used BERT-based-uncased tokenizer to tokenize the text, bert-based-uncased model as our text encoder, and used average pooling over the output of the encoder, a two-layer MLP with a 128 hidden size and $tanh$ as its activation function to predict the labels. The max sentence length is set as 256. We remained the first 256 tokens for sentences that exceed the limit. The learning rate is 1e-5 for BERT encoder, 1e-3 for MLP. For $\alpha$ in the beta distribution, generally, when labeled data is fewer than 100 per class, $\alpha$ is set as 2 or 16, as larger $\alpha$ is more likely to generate $\lambda$ around 0.5, thus creating ``newer" data as data augmentations; when labeled data is more than 200 per class, $\alpha$ is set to 0.2 or 0.4, as smaller $\alpha$ is more likely to generate $\lambda$ around 0.1, thus creating ``similar" data as adding noise regularization.  

For \textbf{TMix}, we only utilize the labeled dataset as the settings in Bert baseline, and set the batch size as 8.
In \textbf{MixText},  we utilize both labeled data and unlabeled data for training using the same settings as in UDA. We set $K = 2$, i.e., for each unlabeled data we perform two augmentations, specifically German and Russian. The batch size is 4 for labeled data and 8 for unlabeled data. 0.5 is used as a starting point to tune temperature $T$. In our experiments, we set 0.3 for AG News, 0.5 for DBpedia and Yahoo! Answer, and 1 for IMDB.

\begin{figure*}[ht]
\centering
\includegraphics[width=2.1\columnwidth]{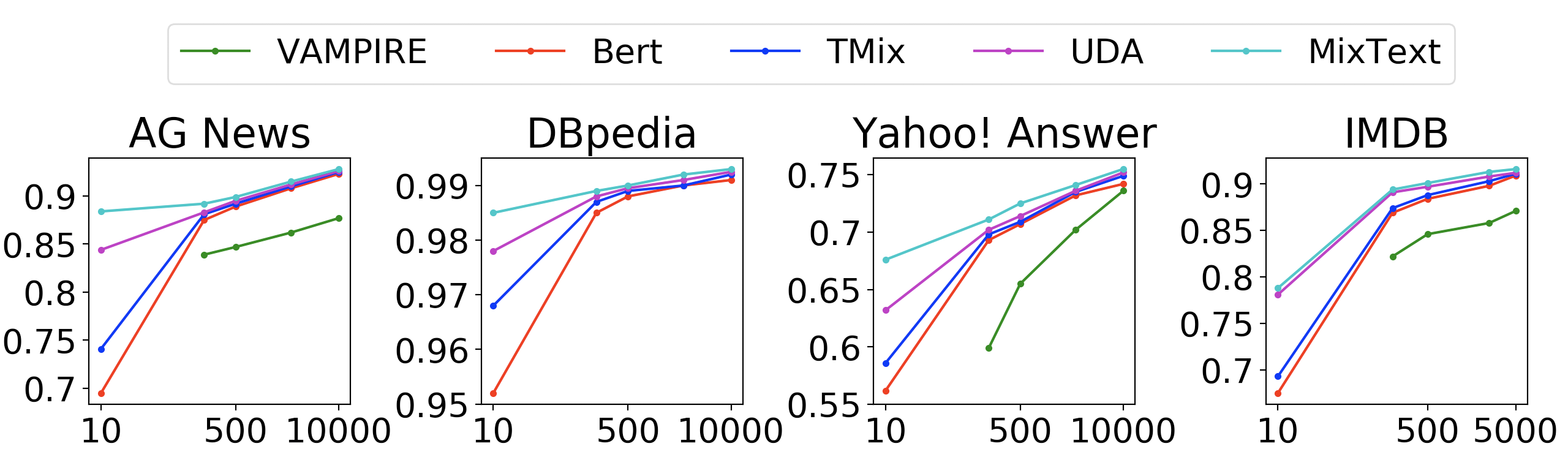}
\caption{Performance (test accuracy (\%)) on AG News, DBpedia, Yahoo! Answer and IMDB with 5000 unlabeled data and varying number of labeled data per class for each model.}\label{fig:main_figure}
\end{figure*}

\subsection{Results}
We evaluated our baselines and proposed methods using accuracy with 5000 unlabeled data and with different amount of labeled data per class ranging from 10 to 10000 (5000 for IMDB).

\subsubsection{Varying the Number of Labeled Data} 
The results on different text classification datasets are shown in Table~\ref{tab:main_result} and Figure~\ref{fig:main_figure}. All transformer based models (BERT, TMix, UDA and MixText) showed better performance compared to VAMPIRE since larger models were adopted. 
TMix outperformed BERT, especially when labeled data was limited like 10 per class. For instance, model accuracy improved from 69.5\% to 74.1\% on AG News with 10 labeled data, demonstrating the effectiveness of TMix. 
When unlabeled data was introduced in UDA, it outperformed TMix such as from 58.6\% to 63.2\% on Yahoo! with 10 labeled data, because more data was used and consistency regularization loss was added. Our proposed MixText consistently demonstrated the best performances when compared to different baseline models across four datasets, as MixText not only incorporated unlabeled data and utilized implicit relations between both labeled data and unlabeled data via TMix, but also had better label guessing on unlabeled data through weighted average among augmented and original sentences.

\begin{figure}[ht!]
\centering
\includegraphics[width=1.0\columnwidth]{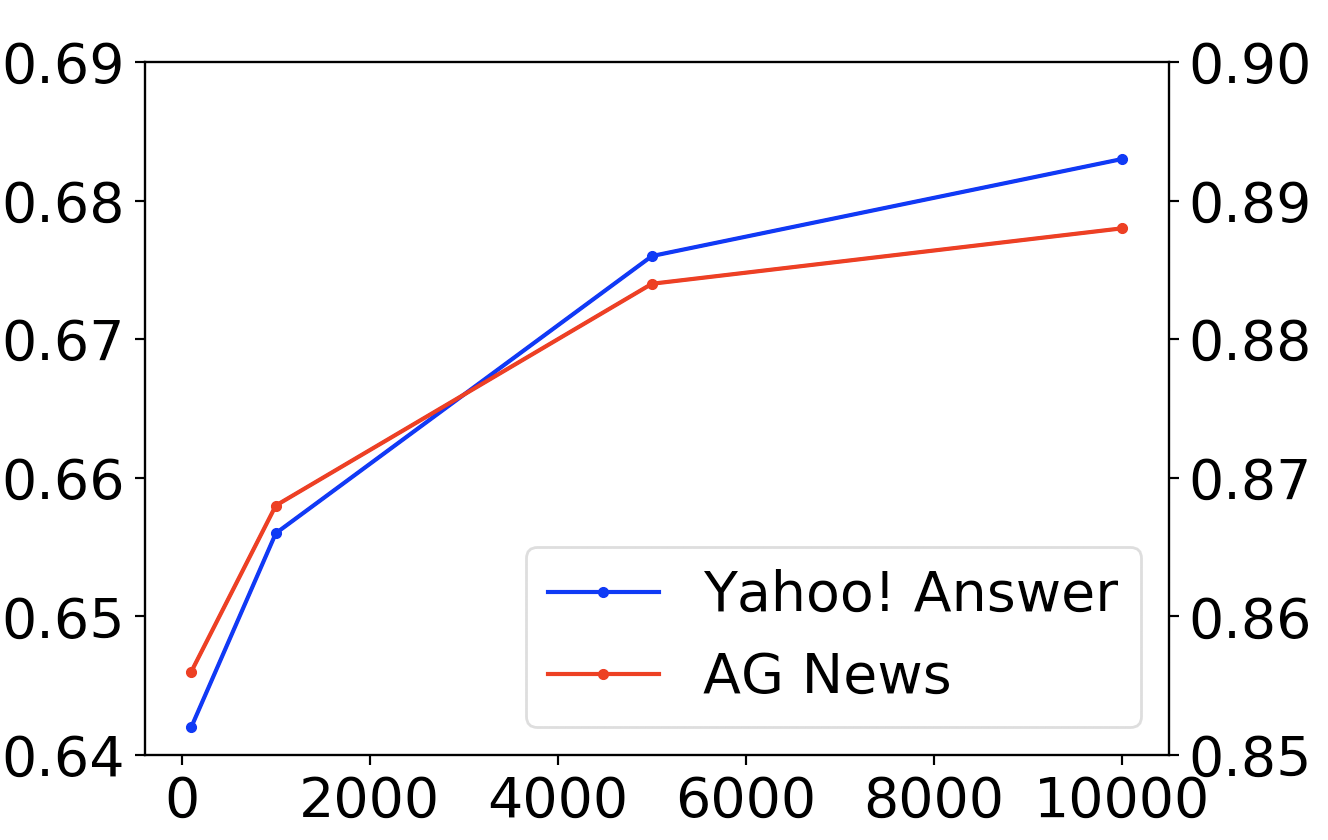}
\caption{Performance (test accuracy (\%)) on AG News ($y$ axis on the right) and Yahoo! Answer ($y$ axis on the left) with 10 labeled data and varying number of unlabeled data per class for MixText. }\label{fig:unlabeled_data}
\end{figure}

\subsubsection{Varying the Number of Unlabeled Data} We also conducted experiments to test our model performances with 10 labeled data and different amount of unlabeled data (from 0 to 10000) on AG News and Yahoo! Answer, shown in Figure~\ref{fig:unlabeled_data}. With more unlabeled data, the accuracy became much higher on both AG News and Yahoo! Answer, which further validated the effectiveness of the usage of unlabeled data.

\subsubsection{Loss on Development Set} 
\begin{figure}[ht]
\centering
\includegraphics[width=1\columnwidth]{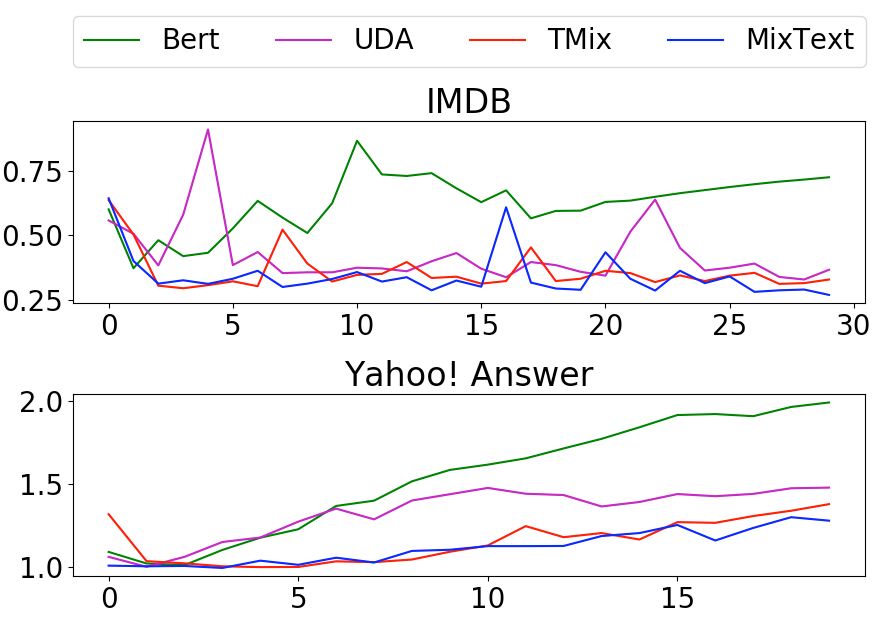}
\caption{Loss on development set on IMDB and Yahoo! Answer in each epoch while training with 200 labeled data and 5000 unlabeled data per class.}\label{fig:loss}
\end{figure}

To explore whether our methods can avoid overfitting when given limited labeled data, we plotted the losses on development set during the training on IMDB and Yahoo! Answer with 200 labeled data per class in Figure~\ref{fig:loss}.  We found that the loss on development sets tends to increase a lot in around 10 epochs for Bert, indicating that the model overfitted on training set. Although UDA can alleviate the overfitting problems with consistency regularization, TMix and MixText showed more stable trends and lower loss consistently. The loss curve for TMix also indicated that it can help solving overfitting problems even without extra data.

\begin{table}[t]
\centering
\begin{tabular}{cc}
\hline
\textbf{Mixup Layers Set} & \textbf{Accuracy(\%)}   \\ \hline 
$\emptyset$ &69.5 \\
 \{0,1,2\} &69.3 \\ 
  \{3,4\} &70.4 \\ 
   \{6,7,9\} &71.9 \\ 
    \{7,9,12\} &\textbf{74.1} \\ 
    \{6,7,9,12\} &72.2 \\ 
     \{3,4,6,7,9,12\} &71.6 \\
\hline 

\end{tabular}\caption{Performance (test accuracy (\%)) on AG News with 10 labeled data per class with different mixup layers set for TMix. $\emptyset$ means no mixup. }\label{tab:layerset}
\end{table}

\begin{table}[ht]
\centering
\begin{tabular}{cc}
\hline
\textbf{Model} & \textbf{Accuracy(\%)}   \\ \hline 
MixText &\textbf{67.6} \\ \hline
 - weighted average & 67.1 \\
 - TMix &63.5 \\ 
 - unlabeled data &58.6 \\ 
 - all &56.2 \\

\hline 

\end{tabular}\caption{Performance (test accuracy (\%)) on Yahoo! Answer with 10 labeled data and 5000 unlabeled data per class after removing different parts of MixText.}\label{tab:ablation}
\end{table}
\subsection{Ablation Studies} \label{ablation}
We performed ablation studies to show the effectiveness of each component in MixText.

\subsubsection{Different Mix Layer Set in TMix} We explored different mixup layer set $M$ for TMix and the results are shown in Table~\ref{tab:layerset}. 
Based on \cite{jawahar-etal-2019-bert}, the \{3,4,5,6,7,9,12\} are the most informative layers in BERT based model and each of them captures different types of information (e.g., surface, syntactic, or semantic).
We chose to mixup using different subsets of those layers to see which subsets gave the optimal performance.
When no mixup is performed, our model accuracy was 69.5\%. 
If we just mixup at the input and lower layers (\{0, 1, 2\}), there seemed no performance increase.
When doing mixup using different layer sets (e.g., \{3,4\}, or \{6,7,9\}), we found large differences in terms of model performances: \{3,4\} that mainly contains surface information like sentence length does not help text classification a lot, thus showing weaker performance. The 6th layer captures depth of the syntactic tree which also does not help much in classifications. Our model achieved the best performance at \{7, 9, 12\}; this layer subset contains most of syntactic and semantic information such as the sequence of top level constituents in the syntax tree, the object number in main clause,  sensitivity to word order, and the sensitivity to random replacement of a noun/verb. 

\subsubsection{Remove Different Parts from MixText} 
We also measured the performance of MixText by stripping each component each time and displayed the results in Table~\ref{tab:ablation}. We observed the performance drops after removing each part, suggesting that all components in MixText contribute to the final performance. The model performance decreased most significantly after removing unlabeled data which is as expected. Comparing to weighted average prediction for unlabeled data, the decrease from removing TMix was larger, indicating that TMix has the largest impact other than unlabeled data, which also proved the effectiveness of our proposed Text Mixup, an interpolation-based regularization and augmentation technique.
\section{Conclusion}
To alleviate the dependencies of supervised models on labeled data, this work presented a simple but effective semi-supervised learning method, MixText, for text classification, in which we also introduced TMix, an interpolation-based augmentation and regularization technique. Through experiments on four benchmark text classification datasets, we demonstrated the effectiveness of our proposed TMix technique and the Mixup model, which have better testing accuracy and more stable loss trend, compared with current pre-training and fine-tuning models and other state-of-the-art semi-supervised learning methods. For future direction, we plan to explore the effectiveness of MixText in other NLP tasks such as sequential labeling tasks and other real-world scenarios with limited labeled data.

\section*{Acknowledgement}
We would like to thank the anonymous reviewers for their helpful comments,
and Chao Zhang for his early feedback.
We gratefully acknowledge the support of NVIDIA Corporation with the donation of the Titan V GPU used for this research.
DY is supported in part by a grant from Google.
\bibliography{acl2020}
\bibliographystyle{acl_natbib}

\end{document}